\documentclass[11pt]{article}

\usepackage[preprint]{acl}

\usepackage{times}
\usepackage{latexsym}

\usepackage[T1]{fontenc}

\usepackage[utf8]{inputenc}

\usepackage{microtype}

\usepackage{inconsolata}
\usepackage{booktabs}   
\usepackage{array} 
\usepackage{xcolor}
\usepackage{colortbl}

\usepackage{graphicx}
\usepackage{amsmath}
\usepackage{polyglossia}
\setotherlanguage{bengali}
\usepackage{listings}
\usepackage{multirow}
\definecolor{cehheader}{HTML}{D6E2F0}      
\definecolor{cehrow}{HTML}{F5F8FC}         
\definecolor{cehaccent}{HTML}{FBF3E4}      
\definecolor{cehgreenfill}{HTML}{E6F0E4}   
\definecolor{cehgreenline}{HTML}{9BC49A}   
\definecolor{cehbox}{HTML}{E4ECF7}         

\lstdefinestyle{jsonstyle}{
  basicstyle=\ttfamily\footnotesize,
  breaklines=true,
  frame=single,
  rulecolor=\color{gray!40},
  backgroundcolor=\color{gray!5},
  columns=fullflexible,
  keepspaces=true,
  showstringspaces=false,
  literate=%
    {গ}{{\textbengali{গ}}}1
    {গন}{{\textbengali{গন}}}1
}

\usepackage{polyglossia}
\setmainlanguage{english}
\setotherlanguage{bengali}

\newfontfamily\bengalifont[Path=./, Script=Bengali]{NotoSerifBengali-Regular.ttf}

\newfontfamily\bengalifonttt[Path=./, Script=Bengali]{NotoSerifBengali-Regular.ttf}
\title{When a Name Is Not a Name: A Benchmark Dataset and Distilled Reasoning for Culturally Entangled Bangla Homographs in Low-Resource LLMs}
\author{
\textbf{Md. Asaduzzaman Shuvo}
\\
\\
United International University, Bangladesh 
\\
\small{
 \textbf{Emails:} \texttt{ashuvo221104@bscse.uiu.ac.bd}} }

\begin{document}
\maketitle
\begin{abstract}
Many Bangla words are at once personal names and culturally loaded common
nouns: \textbengali{মায়া} (\textit{Maya}) is both a girl's name and a word
for affectionate compassion. Choosing the right reading demands cultural
knowledge that is scarce in the pretraining data of modern language models.
We introduce \textbf{Culturally Entangled Homograph (CEH)} disambiguation
and build a Bangla benchmark of 1,516 expert-verified sentences (3,032
labelled occurrences) in which one word appears twice with two distinct
readings, each labelled with a culturally grounded category and an
explanation of the reasoning behind it. Across open- and closed-source
models, we find a systematic \emph{dominant-meaning bias}: models default
to the common-noun sense and overlook the name. A Bangla-specific model
fails under every prompting regime we test, showing that language-specific
pretraining alone does not confer cultural grounding. We further show that
contrastive chain-of-thought prompting can sharply reduce this bias without
training, and that distilling cultural explanations teaches small (1--3B)
models to reason toward the correct reading rather than memorise labels,
cutting dominant-meaning bias from as high as 100\% to under 5\% and
turning the failed Bangla-specific model into our strongest system. We
release the code and the dataset.\footnote{Dataset and code are available at \url{https://github.com/ashuvo25/BanglaCEH}.}
\end{abstract}

\section{Introduction}

Word-sense ambiguity is a long-standing challenge in NLP, but a
particularly difficult and understudied variant arises when a single word
functions simultaneously as a personal name and as a culturally
loaded common noun. In Bangla this is pervasive: parents routinely name
children after words denoting prized emotions, virtues, or spiritual
states, so one surface form carries both an individuating (name) reading
and an abstract (concept) reading.
\begin{figure}[t]
    \centering
    \includegraphics[width=\columnwidth]{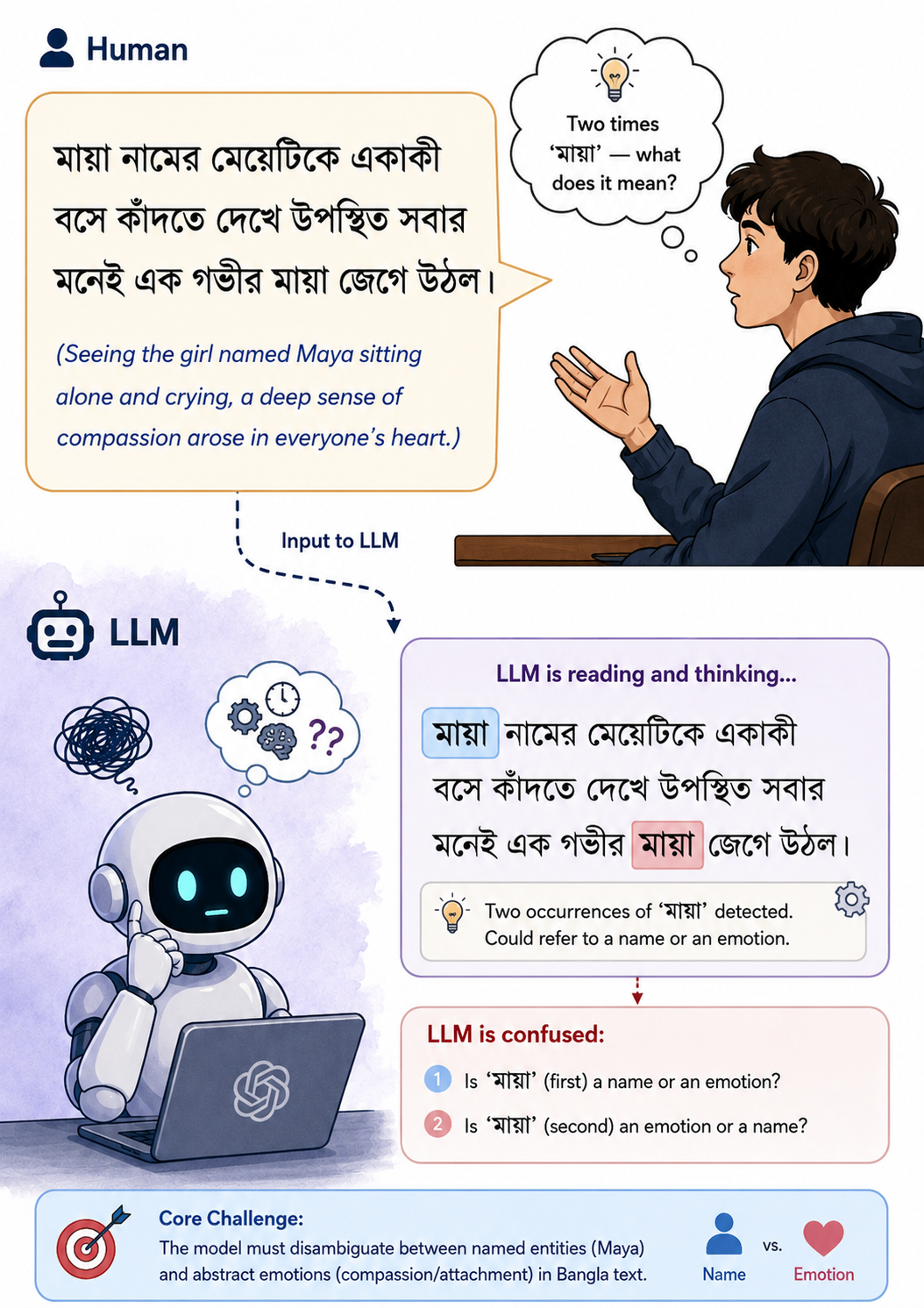}
    \caption{The Culturally Entangled Homograph (CEH) task.
    \textbengali{মায়া} (\textit{Maya}) appears twice in one Bangla
    sentence first as a \emph{name}, then as the \emph{emotion} it is
    named after and the model must disambiguate the two using cultural
    knowledge alone.}
    \label{fig:teaser}
\end{figure}
The word \textbengali{মায়া} (\textit{Maya}) is at once a common girl's
name and a word for deep affectionate compassion;
\textbengali{আরিফ} (\textit{Arif}) is both a boy's name and, in Sufi
theology, one who possesses intuitive knowledge of the divine. Resolving
the intended reading demands cultural and pragmatic knowledge beyond
surface lexical statistics.

We term this phenomenon the \textbf{Culturally Entangled Homograph (CEH)}
and argue it is a sharp diagnostic of genuine cultural grounding. Unlike
conventional word-sense disambiguation, a CEH cannot be resolved by
frequency or distributional cues: both readings coexist in the same
cultural space, and disambiguation hinges on recognising when a form
\emph{names} a person versus \emph{invokes} the concept it was drawn
from making CEH especially demanding for models trained on
high-resource, Western-centric corpora.

LLMs are known to encode a Western-dominance bias and to struggle with
low-resource cultures \citep{belay2025culemo,cheng2025entangled};
language-specific competence does not entail cultural competence, and
in-language prompting alone does not supply cultural context
\citep{belay2025culemo}. For Bangla, recent work has produced capable
models and benchmarks \citep{nahin2025titullms,raihan2025tigerllm,
joy2025bnmmlu}, but evaluation has centred on knowledge, reasoning, and
classification, leaving cultural grounding in everyday lexical usage
largely unexamined. Related studies show that fluent Bangla output can
mask systematic register failures \citep{shuvo2026polite} and persistent
hallucination \citep{ahmed2026benhalluval} symptoms of shallow rather
than grounded understanding.

We introduce the CEH task and build a benchmark of Bangla sentences in
which one word appears twice with two distinct readings, each annotated
with one of six culturally grounded categories and an explanation of the
underlying cultural reasoning. Evaluating open- and closed-source models,
we uncover a systematic \emph{dominant-meaning bias}: models default to
the common-noun reading and overlook the personal-name reading
(Figure~\ref{fig:teaser}). A Bangla-specific model fails under every
regime we test, confirming that language-specific pretraining alone does
not confer cultural grounding. We further show that contrastive
chain-of-thought prompting can sharply reduce this bias without training,
though its effect is strongly model-dependent, and that distilling
cultural reasoning into small models teaches them to reason toward the
correct reading rather than memorise labels.

\textbf{Contributions.}
(i) We formalise CEH disambiguation, a task isolating cultural grounding
from surface lexical competence in a low-resource language. (ii) We
release a Bangla CEH benchmark with token-level cultural categories and
cultural-reasoning explanations. (iii) We evaluate open- and closed-source
models under zero-shot, few-shot, contrastive chain-of-thought, and
knowledge-distillation regimes, revealing a consistent dominant-meaning
bias and the failure of a Bangla-specific model across all settings.
(iv) We show contrastive reasoning and distilled supervision mitigate
this bias, and analyse when and why they succeed or fail.
\section{Related Work}

Word-sense disambiguation (WSD) is a long-standing NLP problem
\citep{navigli2010babelnet,raganato2017wsd}, and remains especially hard
for morphologically rich, low-resource languages where annotated corpora
are scarce \citep{habtamu2024wsd,al2024wsdsurvey}; recent work has extended
WSD evaluation to under-resourced settings through hybrid sense annotation
\citep{senwich2025} and dedicated Arabic and Urdu resources
\citep{arabicnlu2024,saeed2019urdu}, while studies probing whether large
language models truly grasp word senses report that surface fluency often
masks shallow lexical understanding
\citep{meconi2025wordsenses,ortega2023ambiguity}. Our CEH task differs from
classical WSD in that the competing readings a personal name versus the
culturally loaded concept it derives from are not lexically distinct
senses but culturally entangled ones, requiring pragmatic and cultural
knowledge rather than dictionary sense inventories. This connects our work
to a rapidly growing literature on the cultural competence of LLMs, which
finds that models encode a Western-dominance bias and underperform on the
values and everyday knowledge of low-resource cultures
\citep{belay2025culemo,cheng2025entangled}, and that language-specific
ability does not guarantee cultural grounding \citep{belay2025culemo}. For
Bangla specifically, recent efforts have produced dedicated language models
and knowledge benchmarks \citep{nahin2025titullms,raihan2025tigerllm,
joy2025bnmmlu}, yet cultural grounding in everyday lexical usage remains
underexplored, and fluent Bangla generation can still hide systematic
register and factuality failures
\citep{shuvo2026polite,ahmed2026benhalluval}. Finally, our mitigation
strategy builds on chain-of-thought prompting \citep{wei2022cot} and on
knowledge distillation of reasoning into smaller models
\citep{magister2023teaching,hsieh2023distilling}, where a teacher model's
rationales are distilled into a compact student; unlike prior distillation
work that targets mathematical or commonsense reasoning, we distil
\emph{cultural} reasoning, teaching small models to justify why a given
occurrence is a name or a concept rather than merely to reproduce the
label.

\section{Dataset Construction}
\label{sec:dataset}

\textbf{Task formulation.} Each instance in our benchmark is a single
Bangla sentence in which one word form appears \emph{twice} with two
distinct readings. Given the sentence and the target word, a model must
assign each of the two occurrences one of six culturally grounded
categories Name, Concept, Emotion, Emotional State, Collective State, and
Spiritual and justify each assignment. Crucially, the two
readings are not arbitrary lexical senses but \emph{culturally entangled}
ones: the same form functions once as a personal name and once as the
emotion, concept, or spiritual state from which that name is
conventionally drawn.\\
\textbf{Word selection.} We do not rely on a fixed name registry or
lexical database. Instead, a native Bangla speaker manually curated
words such as \textbengali{মায়া} (\textit{Maya}), \textbengali{আশা}
(\textit{Asha}), and \textbengali{গগন} (\textit{Gagan}) that are both
common Bangla given names and culturally loaded common nouns. This
expert-driven selection deliberately targets forms with a genuine dual
reading, which a generic name list would not isolate, at the cost of
coverage reflecting a single annotator's naming knowledge Section
(\ref{sec:limitations}).\\
\textbf{Multi-model generation.} For each selected word, we generated
sentences and accompanying cultural explanations using three
state-of-the-art models as complementary generators: Claude Fable 5, Kimi
K3, and Gemini 3.5 Flash, producing 690, 480, and 344 instances
respectively. Multiple generators reduce the stylistic and distributional
bias of a single teacher and yield greater lexical and structural
diversity.\\
\definecolor{cehgreenfill}{HTML}{E6F0E4}   
\definecolor{cehgreenline}{HTML}{9BC49A}   
\begin{table}[!h]
\centering
\small
\setlength{\arrayrulewidth}{0.3pt}
\setlength{\fboxrule}{0.8pt}
\fcolorbox{cehgreenline}{cehgreenfill}{%
\begin{tabular}{@{}>{\raggedright\arraybackslash}p{0.24\columnwidth}
                  >{\raggedright\arraybackslash}p{0.62\columnwidth}@{}}
\toprule
\textbf{Word} & \textbengali{গগন} (\textit{Gagan}) \\
\textbf{Category} & \textsc{Name} $\leftrightarrow$ \textsc{Concept} \\
\midrule
\textbf{Sentence} &
\textbengali{সমুদ্রসৈকতে সূর্যাস্ত দেখতে দেখতে \textbf{গগন}-র মনে হলো,
প্রাচীন শ্লোকে '\textbf{গগন}' দিয়ে ঠিক এই আকাশ; নভোমণ্ডল-কেই বোঝানো হতো।} \\
\textbf{Gloss} &
\textit{Watching the sunset, it struck \textbf{Gagan} that ancient verse
used `\textbf{Gagan}' for exactly this the celestial firmament.} \\
\midrule
\textbf{Occ.\ 1} & \textsc{Name}   a person, the grammatical subject \\
\textbf{Occ.\ 2} & \textsc{Concept}   the celestial firmament \\
\midrule
\textbf{Cultural note} &
Bengali naming turns words for nature, deities, and virtues into personal
names; \textbengali{গগন} is thus once a living identity and once a cultural
concept. \\
\bottomrule
\end{tabular}}
\caption{An example CEH instance. The same form \textbengali{গগন}
(\textit{Gagan}) appears twice with a name reading and a concept reading.
The full bilingual schema is in Appendix~\ref{app:schema}.}
\label{tab:example}
\end{table}
\textbf{Cross-model and human verification.} We applied a two-stage
verification protocol. First, outputs were cross-checked round-robin by a
different model (Claude verified by Kimi, Kimi by Gemini, Gemini by
Claude), flagging inconsistent labels or explanations without any model
auditing its own output. Second, every instance was manually reviewed by two
native Bangla-speakers (profile in appendix \ref{appendix:annotators}), who corrected labels, explanations, and
culturally inaccurate reasoning. Human verification is the final
authority; the cross-model stage only surfaces candidate errors for the
reviewer.\\
\begin{table}[!h]
\centering
\small
\setlength{\tabcolsep}{3pt}
\begin{tabular}{@{}l@{\hspace{3pt}}r@{\hspace{10pt}}l@{\hspace{3pt}}r@{}}
\toprule
\textbf{Category} & \textbf{Count} & \textbf{Label} & \textbf{Count} \\
\midrule
Name\,$\leftrightarrow$\,Concept    & 795 & Name             & 1{,}153 \\
Concept\,$\leftrightarrow$\,Emotion & 193 & Concept          & 985 \\
Name\,$\leftrightarrow$\,Spiritual  & 179 & Emotional State  & 345 \\
Name\,$\leftrightarrow$\,Emotion    & 178 & Collective State & 202 \\
Emotion\,$\leftrightarrow$\,State   & 171 & Spiritual        & 187 \\
                                    &     & Emotion          & 160 \\
\midrule
\textbf{Instances} & \textbf{1{,}516} & \textbf{Labels} & \textbf{3{,}032} \\
\bottomrule
\end{tabular}
\caption{CEH statistics entanglement categories (left) and
token-level label distribution (right).}
\label{tab:stats}
\end{table}

\textbf{Annotations.} Each verified instance contains token-level labels
for both occurrences, a natural-language explanation of why each
occurrence carries its label, and a concise cultural-entanglement note
describing the naming convention that links the two readings. These
explanations form the supervision signal for our knowledge-distillation
\begin{figure*}[t]
    \centering
    \includegraphics[width=\textwidth]{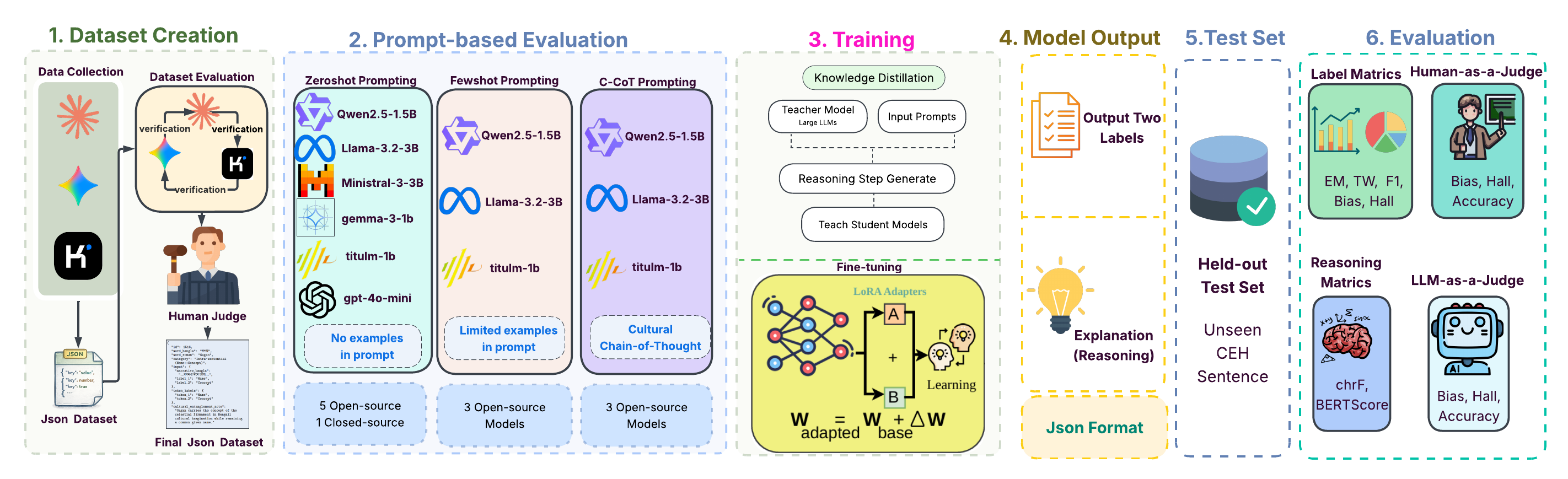}
    \caption{Overview of our methodology, spanning dataset creation,
    prompt-based evaluation (zero-shot, few-shot, C-CoT), knowledge-distillation
    fine-tuning with LoRA, and evaluation on held-out CEH sentences using
    automatic, human, and LLM-based judges.}
    \label{fig:methodology}
\end{figure*}
experiments Section (\ref{sec:method}). An example instance is shown in
Table~\ref{tab:example}; the complete field schema and a raw JSON record
are provided in Appendix~\ref{app:schema}.\\
\textbf{Statistics.} The final benchmark contains 1{,}516 verified
instances, each contributing two labelled occurrences (3{,}032 labelled
tokens). Table~\ref{tab:stats} reports the distribution over entanglement
categories and over the six labels. Name and Concept are
the most frequent labels, reflecting the prominence of name--concept
entanglement in Bangla naming practice, while Emotion and Spiritual form a rarer long tail that we find models handle least reliably in  Section (\ref{sec:results}).

\section{Methodology}
\label{sec:method}
We evaluate the CEH task under four regimes of increasing supervision:
zero-shot prompting, few-shot prompting, Cultural Chain-of-Thought
(C-CoT) prompting, and knowledge-distillation (KD) fine-tuning. The first two
establish how much cultural grounding models possess without task-specific
adaptation; the latter two are our proposed interventions for mitigating
the dominant-meaning bias. All regimes are evaluated on the same held-out
test set to enable direct comparison. We split the benchmark into 78\%
training, 10\% validation, and 12\% test partitions with a fixed random
seed. Figure~\ref{fig:methodology} provides an overview of the full
pipeline.
\subsection{Prompting Baselines}

\paragraph{Zero-shot.} The model receives the sentence and an instruction
to assign each of the two occurrences of the target word a label from the
six categories, with no examples. This measures a model's out-of-the-box
cultural grounding.
\textbf{Few-shot.} We prepend a small number of labelled in-context
examples before the query, testing whether demonstrations alone without
weight updates help the model recognise the entangled readings.

\subsection{Cultural Chain-of-Thought}

Standard chain-of-thought prompting elicits intermediate reasoning before a
final answer \citep{wei2022cot}, but it does not counteract a model's prior
tendency to collapse both occurrences onto the dominant sense. We introduce
\textbf{Cultural Chain-of-Thought (C-CoT)} (Appendix \ref{app:ccot-prompt}), a contrastive variant that
explicitly forces the model to argue \emph{both} candidate readings for
each occurrence before committing. For every occurrence, the model must (i)
state why the form could be a \textsc{Name}, (ii) state why it could be the
competing cultural category (\textsc{Concept}, \textsc{Emotion},
\textsc{Spiritual}, or \textsc{State}), and (iii) decide which reading the
sentential and cultural context supports. By requiring the suppressed
reading to be articulated before a decision is made, C-CoT directly targets
the dominant-meaning bias rather than merely eliciting free-form rationale.
C-CoT requires no training and is applied at inference time.

\subsection{KD Fine-Tuning}

Our final regime tests whether the cultural reasoning behind each label can
be \emph{taught} to a small model rather than merely prompted. We fine-tune
compact open-source models on the human-verified cultural explanations in
our benchmark, following a knowledge-distillation formulation in which a
student model learns to reproduce the reasoning traces of stronger teacher
models \citep{hsieh2023distilling,magister2023teaching}. Crucially, the
training target is the \emph{full cultural explanation followed by the
label}, not the label alone: the model first generates the reasoning that
justifies each occurrence's reading and only then emits the final labels.
This encourages the model to internalise the pattern of cultural reasoning
rather than memorise a surface mapping from word to label.
We use parameter-efficient fine-tuning via Low-Rank Adaptation
\citep{hu2022lora} in its quantized form, QLoRA \citep{dettmers2023qlora},
which back-propagates gradients through a frozen 4-bit base model into
low-rank adapters, enabling training of billion-parameter models on a
single consumer GPU. Loss is computed only over the target reasoning and
label tokens, with the prompt masked, so that supervision is concentrated
on the cultural reasoning the model must learn to produce. We select the
checkpoint with the lowest validation loss and apply early stopping to
prevent overfitting.

\subsection{Models}

We evaluate a mix of open- and closed-source models spanning a range of
parameter scales. The open-source set comprises Qwen2.5-1.5B, Llama-3.2-3B,
Gemma-3-1B,  Ministral-3B, and the Bangla-specific TituLLM; the last is
included specifically to test whether language-targeted pretraining confers
cultural grounding. We include GPT-4o-mini as a strong closed-source
reference. The prompting baselines are evaluated across all models, while
C-CoT and knowledge-distillation fine-tuning are applied to a representative
subset spanning small general-purpose and Bangla-specific models.

\subsection{Evaluation Metrics}

We evaluate along two tracks. The label track measures
disambiguation accuracy through Exact Match (both occurrences correct), Average Score (Average Score is the percentage of total points earned over the maximum possible points, see Appendix(\ref{app:Average-Score})), together with Macro- and Micro-averaged F1 over the six labels. We additionally report a Hallucination rate the fraction of outputs producing a label outside the valid set and, as our central diagnostic, a Dominant-Bias rate measuring how often a model misses a Name label and defaults to the entangled cultural reading. The
reasoning track evaluates the quality of generated cultural explanations against the gold explanations using chrF \citep{popovic2015chrf} and BERTScore \citep{zhang2020bertscore}. In addition, we assess outputs through a Human-as-a-Judge and LLM-as-a-Judge evaluation.

\section{Result Analysis}
\label{sec:results}

Table~\ref{tab:main} reports performance across all four regimes. We
organise our analysis around four findings: the severity of the
dominant-meaning bias under prompting, the model-dependent behaviour of
C-CoT, the decisive effect of knowledge-distillation fine-tuning, and the
striking reversal of the Bangla-specific model.
\begin{table*}[t]
\centering
\small
\setlength{\tabcolsep}{4.5pt}
\begin{tabular}{@{}ll cccc cc cc@{}}
\toprule
& & \textbf{EM}$\uparrow$ & \textbf{Avg}$\uparrow$
& \textbf{Ma-F1}$\uparrow$ & \textbf{Mi-F1}$\uparrow$
& \textbf{chrF}$\uparrow$ & \textbf{BERT}$\uparrow$
& \textbf{Hall.}$\downarrow$ & \textbf{Bias}$\downarrow$ \\
\midrule
\multicolumn{10}{@{}l}{\textbf{\textit{Open-source}}} \\
\midrule
\multirow{4}{*}{Qwen2.5-1.5B}
 & Zero-shot   & 0.65  & 22.90 & 0.115 & 0.229 & 1.96  & 0.291 & 52.90 & 63.46 \\
 & Few-shot    & 8.39  & 36.45 & 0.227 & 0.365 & 17.70 & 0.301 & 54.84 & 23.08 \\
 & C-CoT       & 30.97 & 52.26 & 0.300 & 0.523 & 24.30 & 0.178 & \textbf{0.00} & 2.88 \\
 & QLoRA-KD    & \textbf{85.16} & \textbf{92.58} & \textbf{0.737} & \textbf{0.926} & \textbf{79.35} & \textbf{0.784} & 1.65 & \textbf{0.00} \\
\midrule
\multirow{4}{*}{Llama-3.2-3B}
 & Zero-shot   & 10.97 & 20.97 & 0.144 & 0.210 & 1.71  & 0.295 & 40.00 & 65.38 \\
 & Few-shot    & 22.58 & 40.00 & 0.239 & 0.400 & 16.20 & 0.289 & 34.19 & 22.12 \\
 & C-CoT       & 15.16 & 15.81 & 0.153 & 0.158 & 27.80 & 0.144 & 43.23 & 75.00 \\
 & QLoRA-KD    & \textbf{82.42} & \textbf{90.38} & \textbf{0.722} & \textbf{0.904} & \textbf{77.04} & \textbf{0.760} & \textbf{1.10} & \textbf{2.80} \\
\midrule
\multirow{2}{*}{Ministral-3B}
 & Zero-shot   & \textbf{9.00}  & \textbf{22.60} & \textbf{0.158} & \textbf{0.226} & 2.19  & 0.251 & 41.30 & 62.50 \\
 & Few-shot    & 5.80  & 20.30 & 0.126 & 0.203 & \textbf{10.92} & \textbf{0.270} & \textbf{30.30} & \textbf{60.60} \\
\midrule
\multirow{2}{*}{gemma-3-1b-it}
 & Zero-shot   & 0.91  & 13.23 & 0.079 & 0.132 & 10.01 & 0.072 & 16.13 & 100.0 \\
 & Few-shot    & \textbf{31.29} & \textbf{54.19} & \textbf{0.339} & \textbf{0.542} & \textbf{17.59} & \textbf{0.285} & \textbf{7.74}  & \textbf{39.50} \\
\midrule
\multirow{4}{*}{titulm-1b}
 & Zero-shot   & 0.11  & 2.33  & 0.005 & 0.006 & 2.33  & 0.032 & 100.0 & 100.0 \\
 & Few-shot    & 0.31  & 33.55 & 0.115 & 0.335 & 5.77  & 0.037 & 100.0 & 40.22 \\
 & C-CoT       & 1.14  & 2.77  & 0.012 & 0.014 & 13.23 & 0.048 & 89.12 & 92.13 \\
 & QLoRA-KD    & \textbf{87.91} & \textbf{92.58} & \textbf{0.758} & \textbf{0.926} & \textbf{79.50} & \textbf{0.790} & \textbf{1.65} & \textbf{4.20} \\
\midrule
\multicolumn{10}{@{}l}{\textbf{\textit{Closed-source}}} \\
\midrule
\multirow{2}{*}{GPT-4o-mini}
 & Zero-shot   & 43.87 & 64.19 & 0.467 & 0.642 & 19.10 & \textbf{0.299} & 0.65 & 16.35 \\
 & Few-shot    & \textbf{49.68} & \textbf{72.26} & \textbf{0.623} & \textbf{0.723} & \textbf{32.60} & 0.296 & \textbf{0.00} & \textbf{2.88} \\
\bottomrule
\end{tabular}
\caption{CEH results across four regimes. $\uparrow$ higher is better;
$\downarrow$ lower is better. EM = Exact Match, Avg = Average Score,
Ma-F1 / Mi-F1 = Macro / Micro F1, BERT = BERTScore-F1, Hall.\ =
Hallucination, Bias = Dominant-Bias (all \%). C-CoT and QLoRA-KD were run on
closed-source prompting reference. For each model, the best value per metric
is in \textbf{bold}.}
\label{tab:main}
\end{table*}

\subsection{Zero-Shot Prompting}
\label{sec:results-zeroshot}

Under zero-shot prompting, every open-source model shows a pronounced
dominant-meaning bias, defaulting to the common-noun reading and
misclassifying the personal-name occurrence. Bias ranges from 62.5\%
( Ministral) to 100\% (Gemma, TituLLM), with exact match below 11\% for all
open models. The failure modes differ:  Ministral and Llama produce valid but
wrong labels, while others collapse entirely TituLLM hallucinates on
100\% of inputs and Gemma shows complete (100\%) bias, indicating the name
reading is not even represented as a candidate.
The closed-source GPT-4o-mini is markedly stronger 43.87\% exact match,
0.65\% hallucination yet retains a non-trivial 16.35\% bias. This
open--closed gap holds across every metric, confirming CEH disambiguation
is genuinely hard: even a capable proprietary model without task-specific
adaptation misreads the name as the entangled concept in roughly one in
six cases. Uniformly low Macro-F1 (0.005--0.467) further shows zero-shot
failure is disproportionate on rarer cultural labels, not merely the
frequent name/concept pairing.

\subsection{Few-Shot Prompting}
\label{sec:results-fewshot}

Adding a few in-context examples improves most models over zero-shot, but
the gains are inconsistent and leave the core bias largely unresolved.
Gemma benefits most dramatically, rising from 0.91\% to 31.29\% exact
match as its dominant-meaning bias falls from 100\% to 39.5\% evidence
that demonstrations can teach a model to represent the name reading it
previously ignored. Qwen and Llama show more modest gains (exact match
0.65\%$\rightarrow$8.39\% and 10.97\%$\rightarrow$22.58\%), with bias
dropping into the low-20\% range for both. Two models resist this trend:
 Ministral \emph{degrades} under few-shot prompting exact match falls from
9.0\% to 5.8\% with bias barely moving (62.5\%$\rightarrow$60.6\%), the
demonstrations disrupting rather than guiding its predictions while
TituLLM remains effectively non-functional, hallucinating on 100\% of
inputs in both regimes despite a nominal rise to 0.31\% exact match,
confirming its failure is one of task representation rather than a
shortage of examples. GPT-4o-mini again leads, reaching 49.68\% exact
match with bias reduced to 2.88\%, its strongest prompting result.
Overall, few-shot prompting narrows but does not close the gap: even the
best open-source configuration (Gemma, 31.29\%) trails GPT-4o-mini by a
wide margin, and dominant-meaning bias remains substantial for every open
model motivating the reasoning- and training-based interventions we
examine next.

\subsection{Cultural Chain-of-Thought}
\label{sec:results-ccot}

C-CoT produces the largest prompting-time improvement we observe, but only
for models able to follow its contrastive structure. On Qwen2.5-1.5B, it
drives dominant-meaning bias from 63.46\% (zero-shot) to 2.88\%,
eliminates hallucination entirely (0\%), and raises exact match to
30.97\% a level no other prompting configuration reaches for this model.
Forcing the model to articulate the suppressed name-reading before
committing counteracts the bias without parameter updates, supporting our
hypothesis that the dominant-meaning failure is one of reasoning order
rather than missing knowledge. The benefit does not transfer uniformly:
on Llama-3.2-3B, C-CoT is counter-productive dominant-meaning bias
\emph{rises} to 75\% and hallucination climbs to 43.23\%, both worse than
the zero- and few-shot baselines. The degradation coincides with a sharp
rise in unparseable outputs: Llama's extended contrastive reasoning often
fails to terminate in the required label format, so otherwise valid
predictions are discarded. C-CoT's effectiveness is thus contingent on
sustaining format adherence over long reasoning chains an important
limitation for smaller and lower-resource models. TituLLM remains largely
non-functional under C-CoT (89.12\% hallucination, 1.14\% exact match);
contrastive reasoning cannot compensate for a model that does not reliably
represent the task. Together, these results show C-CoT can nearly
eliminate dominant-meaning bias when a model can execute it, but not a
universal remedy motivating the training-based approach we examine next.
\subsection{QLoRA-KD Fine-Tuning}
\label{sec:results-finetune}

The QLoRA-KD regime yields the strongest results by a wide margin. All three
fine-tuned models exceed 82\% exact match and reduce dominant-meaning bias
to at most 4.2\% an order-of-magnitude improvement over their best
prompting configuration. Qwen2.5-1.5B reaches 85.16\% exact match with
0\% dominant-meaning bias and 1.65\% wrong predictions,
while Llama-3.2-3B, which C-CoT had degraded, recovers to 82.42\% exact
match and 2.8\% bias. This confirms that the dominant-meaning failure is not
an inherent limitation of these models but a gap that targeted supervision
can close. Crucially, the improvement extends to reasoning quality, not merely label accuracy. For Qwen, BERTScore rises from 0.178 under C-CoT to 0.784 under
QLoRA-KD and chrF from 24.3 to 79.35, indicating that the fine-tuned model
generates cultural explanations closely aligned with the gold reasoning
rather than producing correct labels for spurious reasons. Because the
distillation target pairs each label with its cultural justification, the
model learns the \emph{pattern} of cultural reasoning that licenses a
reading rather than a surface word-to-label mapping precisely the
capability that prompting failed to elicit.

These gains are obtained with parameter-efficient adaptation on models of
only 1--3B parameters. Notably, Qwen2.5 fine-tuned model surpasses the
strongest closed-source prompting baseline (GPT-4o-mini few-shot, 49.68\%
exact match, 2.88\% bias) by a large margin. We stress that this is not a
like-for-like comparison GPT-4o-mini is evaluated only under prompting,
not fine-tuned but it establishes that a small, openly available model,
once taught cultural reasoning, can substantially outperform a much larger
proprietary model prompted on the same task. The residual gap between
Micro- and Macro-F1 ( 0.926 vs.\ 0.737 for Qwen) indicates that errors
now concentrate almost entirely in the rarer cultural labels, which we identify as the primary remaining
challenge (full per-label results and confusion matrices in Appendix~\ref{sec:results-perlabel}). An ablation study is provided in Appendix~\ref{sec:ablation}, confirming the importance of reasoning supervision.

\subsection{Human \& LLM Judge Evaluation}
\label{sec:results-judges}

\begin{table}[!h]
\centering
\small
\setlength{\tabcolsep}{4pt}
\begin{tabular}{@{}l cccc@{}}
\toprule
\textbf{Model} & \textbf{Hall.\%}$\downarrow$ & \textbf{Bias\%}$\downarrow$
& \textbf{Exact\%}$\uparrow$ & \textbf{P\%}$\uparrow$ \\
\midrule
\multicolumn{5}{@{}l}{\textit{Human-as-a-Judge}} \\
Qwen2.5-1.5B  & 1.0 & 0.2 & 84 & 15 \\
Llama-3.2-3B & 1.0 & 2.0 & 82 & 16 \\
titulm-1b & 2.0 & 3.0 & 87 & 10 \\
\midrule
\multicolumn{5}{@{}l}{\textit{LLM-as-a-Judge (GPT-5.4)}} \\
Qwen2.5-1.5B  & 1.0 & 0.3 & 85 & 14 \\
Llama-3.2-3B & 1.1 & 2.5 & 82 & 16 \\
titulm-1b & 2.0 & 3.5 & 87 & 10 \\
\bottomrule
\end{tabular}
\caption{Human and LLM (GPT-5.4) judge evaluation of the fine-tuned
(QLoRA-KD) models. $\uparrow$ higher is better; $\downarrow$ lower is
better and P = Partial.}
\label{tab:judges}
\end{table}

To verify that the fine-tuned models reach correct labels through culturally
sound reasoning, we evaluate their outputs with two-stage independent judges: two human judges (native Bangla speakers) and an LLM judge GPT-5.4 (the template is in Appendix \ref{app:judge-prompt}). Each judge scores every output for hallucination, dominant-meaning bias, and
exact- and partial-match accuracy; results are reported in
Table~\ref{tab:judges}. The two judges agree closely, differing by at most
1 point on any metric, which both corroborates our automatic evaluation
and supports the use of an LLM judge as a scalable proxy for human assessment.
Consistent with the automatic results, all three fine-tuned models exhibit
near-zero hallucination and dominant-meaning bias under both judges, and
TituLLM attains the highest exact-match rating (87\%), confirming that its
post-distillation predictions are not only accurate but judged culturally
faithful by a human expert.

\section{Discussion}
\label{sec:discussion}

Our results separate two mechanisms for resolving culturally entangled
homographs: surfacing knowledge already latent in a model's parameters, and
acquiring knowledge it lacks. Prompting, including contrastive C-CoT, engages
only the former it helps only insofar as cultural grounding is already
present and can be reorganised through reasoning order, which is why C-CoT
nearly eliminates the dominant-meaning bias on Qwen yet leaves TituLLM, which
lacks such grounding, effectively unchanged. Knowledge-distillation
fine-tuning engages the latter: supervising on cultural explanations rather
than labels alone induces the reasoning that connects a name to its source
concept, so predictions follow from reasoning rather than memorised surface
mappings. Under this regime accuracy and reasoning quality improve together
rather than in isolation the models do not merely predict the right label
but justify it with culturally faithful explanations, a coupling our human
and LLM judges independently corroborate.

These gains proved sensitive to how the adaptation was configured. Because
the benchmark is small relative to adapter capacity, an initial
higher-capacity setting overfit within a few epochs; reducing the adapter
rank and learning rate and strengthening regularisation delayed overfitting
and lowered the validation minimum, and combined with early stopping this
configuration transferred unchanged across all three models, indicating that
the recipe is not narrowly tuned to a single architecture
(Appendix~\ref{app:hyperparams}).

The most consequential result is the reversal of the Bangla-specific model:
non-functional under every prompting regime, yet best-in-class after
distillation. This suggests that monolingual pretraining supplies a useful
substrate more faithful tokenisation of Bangla script and denser exposure
to the relevant names and morphology but not task competence itself; the
substrate instead allows limited reasoning supervision can be absorbed
far more efficiently than by a general model with a weaker lexical prior. For
low-resource, culturally rich languages that lack large instruction-tuned
systems, this points to a reproducible recipe: pair a compact
language-specific model with a small set of cultural reasoning traces and
adapt it with parameter-efficient fine-tuning. Although our experiments focus on Bangla, the proposed framework is language-agnostic and could be evaluated on other languages with similar name--concept entanglement.

Two limitations qualify these conclusions. First, the benefit of C-CoT is
model-dependent: it helps only when a model can sustain the required output
format across long reasoning chains, and degrades performance where it
cannot, as observed for Llama. Second, the persistent gap between Micro- and
Macro-averaged F1 after fine-tuning shows that the rarer cultural labels
remain the hardest, locating the current bottleneck in long-tail data
scarcity rather than in the method, and marking targeted expansion of these
categories as the clearest direction for future work.
\section{Conclusion}
\label{sec:conclusion}

We introduced Culturally Entangled Homograph (CEH) disambiguation, a task in
which the same Bangla word must be read once as a personal name and once as
the cultural concept it is drawn from, and released a benchmark of annotated
instances paired with cultural-reasoning explanations. Across open- and
closed-source models we found a systematic dominant-meaning bias models
default to the common-noun sense and overlook the name and showed that
language-specific pretraining alone does not fix it: the Bangla-specific model
failed under every prompting regime. Yet the same model became the strongest
system once fine-tuned on distilled cultural reasoning, reaching near-perfect
disambiguation with the bias eliminated. Our results suggest that cultural grounding is not conferred by monolingual pretraining alone, but can be improved through supervision with reasoning-annotated examples. This provides a practical approach for improving cultural disambiguation in low-resource languages. Evaluating its applicability to other languages with similar name--concept entanglement remains an important direction for future work.

\newpage
\section*{Limitations}
\label{sec:limitations}

Our benchmark has several limitations. First, word selection was performed
manually by a single native Bangla speaker rather than drawn from a name
registry or lexical database; while this deliberately targets forms with a
genuine dual reading, it means coverage reflects one annotator's naming
knowledge and may under-represent regional or less common names. Second,
the sentences and cultural explanations were initially produced by large
language models and subsequently verified by two native Bangla-speaking
experts. Although every instance was manually reviewed, model-generated
text may still exhibit subtle stylistic regularities that a fully
human-authored corpus would not. Third, although every instance was independently reviewed by two native Bangla-speaking annotators and the resulting inter-annotator agreement was high (Cohen's $\kappa = 0.95$), the annotation process still relied on a relatively small annotator pool. While disagreements were resolved through discussion using a shared annotation guideline, involving more annotators from diverse regional and linguistic backgrounds would further improve the robustness and generalizability of the benchmark. Finally, our fine-tuning experiments use parameter-efficient
adaptation on small models under limited compute, so the reported gains
may not transfer directly to larger models or full-parameter fine-tuning.

\bibliography{custom}
\appendix

\section{Annotator Profile}
\label{appendix:annotators}

The dataset was verified by two native Bangla-speaking annotators with
graduate-level backgrounds in Computer Science and prior experience in
Bangla NLP research. Both annotators independently reviewed the labels and
cultural reasoning associated with each instance. Disagreements were
resolved through discussion using a shared annotation guideline. The
resulting inter-annotator agreement, measured using Cohen's $\kappa$, was
0.95, indicating almost perfect agreement.

\section{Result Analysis Extend}
\subsection{Ablation Study}
\label{sec:ablation}

Table~\ref{tab:ablation} examines the contribution of reasoning-aware
knowledge distillation by comparing zero-shot inference, label-only
distillation, and the full QLoRA-KD framework. Zero-shot prompting performs
poorly on the CEH task, achieving only 0.65\% Exact Match with high
dominant-meaning bias (63.46\%) and hallucination (52.9\%), indicating that
the pretrained model lacks sufficient cultural grounding. Introducing
label-only knowledge distillation substantially improves Exact Match to
41.33\% and Macro-F1 to 48.12\%, demonstrating that supervision from teacher
predictions is beneficial. However, the model still exhibits considerable
dominant-meaning bias (34.56\%) and a high hallucination rate (36.11\%),
suggesting that labels alone are insufficient for learning the underlying
cultural distinctions. In contrast, the full QLoRA-KD framework, which
distills both cultural reasoning and labels, achieves 85.16\% Exact Match and
73.7\% Macro-F1 while reducing dominant-meaning bias to 0\% and
 hallucination rate 1.65\%. These results show that the largest gains
are obtained by supervising the reasoning process rather than the prediction
labels alone. These results show that supervising the reasoning process rather than the prediction labels alone enables substantially more reliable cultural disambiguation.

\begin{table}[!h]
\centering
\scriptsize
\setlength{\tabcolsep}{1.8pt}   
\renewcommand{\arraystretch}{0.9} 
\begin{tabular}{lcccc}
\toprule
\textbf{Method} &
\textbf{EM\%} &
\textbf{Bias\%} &
\textbf{Hall.\%} &
\textbf{Macro-F1\%} \\
\midrule
Zero-shot               & 0.65 & 63.46 & 52.9 & 11.50 \\
KD (Label-only)         & 41.33 & 34.56 & 36.11 & 48.12 \\
\textbf{QLoRA-KD}       & \textbf{85.16} & \textbf{0} & \textbf{1.65} & \textbf{73.7} \\
\bottomrule
\end{tabular}
\caption{Ablation study. $\uparrow$ indicates higher is better and $\downarrow$ indicates lower is better.}
\label{tab:ablation}
\end{table}

\section{Dataset Schema and Example Record}
\label{app:schema}

Each instance in the CEH benchmark is stored as a JSON record with the
fields listed below. The body of the paper (Table~\ref{tab:example}) shows
a condensed view; here we give the complete bilingual schema and a full raw
record.
\definecolor{cehbox}{HTML}{E4ECF7}   

\begin{figure}[!h]
\centering
\small
\fcolorbox{cehheader}{cehbox}{\parbox{0.92\columnwidth}{\ttfamily\footnotesize
\{\\
\hspace*{1em}"id": 1516,\\
\hspace*{1em}"word\_bangla": "\textbengali{গগন}",\\
\hspace*{1em}"word\_roman": "Gagan",\\
\hspace*{1em}"category": "Intra-sentential\\
\hspace*{2em}(Name$\leftrightarrow$Concept)",\\
\hspace*{1em}"input": \{\\
\hspace*{2em}"narrative\_bangla":\\
\hspace*{3em}"\textbengali{...গগন-র মনে হলো...}",\\
\hspace*{2em}"label\_1": "Name",\\
\hspace*{2em}"label\_2": "Concept"\\
\hspace*{1em}\},\\
\hspace*{1em}"token\_labels": \{\\
\hspace*{2em}"token\_1": "Name",\\
\hspace*{2em}"token\_2": "Concept"\\
\hspace*{1em}\},\\
\hspace*{1em}"cultural\_entanglement\_note":\\
\hspace*{2em}"Gagan carries the concept of the\\
\hspace*{2em}celestial firmament in Bengali\\
\hspace*{2em}cultural imagination while remaining\\
\hspace*{2em}a common given name."\\
}}
\caption{A raw CEH benchmark record (condensed). Bangla fields render in
native script; the full schema additionally includes bilingual
justifications, cultural-entanglement paragraphs, and full explanations
omitted here for space.}
\label{fig:example-record}
\end{figure}
\newpage
\section{Cultural Chain-of-Thought Prompt}
\label{app:ccot-prompt}

Figure~\ref{fig:ccot-prompt} shows the full Cultural Chain-of-Thought (C-CoT)
prompt template. Unlike standard chain-of-thought, it forces the model to
argue \emph{both} the name and the competing cultural reading for each
occurrence before committing to a label, directly targeting the
dominant-meaning bias.

\begin{figure}[h]
\centering
\small
\fcolorbox{cehheader}{cehrow}{%
\parbox{0.92\columnwidth}{\ttfamily\footnotesize
\textbf{[System]}\\[2pt]
You are an expert in Bangla cultural linguistics. Some Bangla words are
simultaneously a person's NAME and a culturally meaningful concept, emotion,
or spiritual state. Explain WHY each occurrence carries its meaning, then give
the final labels.\\[6pt]
\textbf{[User]}\\[2pt]
The Bangla word `\textbengali{গগন}' (Gagan) appears TWICE in this sentence
with two different meanings.\\[4pt]
Sentence: \{\textit{narrative}\}\\[4pt]
Reason contrastively for EACH occurrence:\\[2pt]
For Occurrence 1:\\
- Argument A: why it could be a Name.\\
- Argument B: why it could be an Emotion / Concept / Spiritual / State.\\
- Decision: which reading the context supports.\\[2pt]
For Occurrence 2:\\
- Argument A: why it could be a Name.\\
- Argument B: why it could be an Emotion / Concept / Spiritual / State.\\
- Decision: which reading the context supports.\\[4pt]
After reasoning, end with EXACTLY this block:\\[2pt]
FINAL:\\
Occurrence 1: \textless label\textgreater\\
Occurrence 2: \textless label\textgreater\\[4pt]
Valid labels: Name, Emotion, Concept, Spiritual, Emotional State,
Collective State
}}
\caption{The Cultural Chain-of-Thought (C-CoT) prompt template. The model must
articulate a Name argument and a competing cultural-category argument for each
occurrence before producing the final labels, counteracting the
dominant-meaning bias.}
\label{fig:ccot-prompt}
\end{figure}
\section{LLM-as-a-Judge Prompt}
\label{app:judge-prompt}

Figure~\ref{fig:judge-prompt} shows the prompt given to the LLM judge
(GPT-5.4) for evaluating the fine-tuned models. To mitigate known judge
biases \citep{zheng2023judging,thakur2024judging}, the judge model is
distinct from all models under evaluation, and the human annotators were given
an identical rubric to ensure the two judges measure the same quantities.

\begin{figure}[h]
\centering
\small
\fcolorbox{cehheader}{cehrow}{%
\parbox{0.92\columnwidth}{\ttfamily\footnotesize
You are an expert evaluator of Bangla cultural linguistics, judging a model's
performance on the Culturally Entangled Homograph (CEH) task, where one Bangla
word appears twice in a sentence with two different meanings (e.g., once as a
NAME and once as a CONCEPT, EMOTION, SPIRITUAL, or STATE reading). You are
given a batch of items, each with the sentence, gold labels, gold reasoning,
and the model output.\\[6pt]
\textbf{Valid labels:} Name, Emotion, Concept, Spiritual, Emotional State,
Collective State.\\[6pt]
\textbf{Per-item judgement:}\\
- \texttt{match\_1}: does the model's Occurrence-1 label equal the gold label?\\
- \texttt{match\_2}: does the model's Occurrence-2 label equal the gold label?\\
- \texttt{bias}: did the model label a gold NAME occurrence as a non-name
(cultural) reading?\\
- \texttt{hallucinated}: did the model output a label outside the valid set,
or fail to produce a parseable label?\\[6pt]
\textbf{Aggregation (over N items):}\\
Exact\% = 100 $\times$ (match\_1 AND match\_2) / N\\
Partial\% = 100 $\times$ (exactly one match) / N\\
Bias\% = 100 $\times$ (bias) / (items with a gold Name)\\
Hall.\% = 100 $\times$ (hallucinated) / N\\[6pt]
\textbf{Output:} return ONLY a JSON object, rounded to one decimal  \\
\{"model", "hallucination\_pct", "bias\_pct", "exact\_pct", "partial\_pct"\}
}}
\caption{The LLM-as-a-judge prompt (GPT-5.4). The judge scores each output for
label match, dominant-meaning bias, and hallucination, then aggregates into
the percentage metrics reported in Table~\ref{tab:judges}.}
\label{fig:judge-prompt}
\end{figure}

\section{Formula}
\subsection{Average Score}
\label{app:Average-Score}
Each CEH instance contains two target occurrences. A prediction receives
one point for each correctly classified occurrence (maximum two points per
instance). The Average Score is computed as

\[
\text{Average Score} =
\frac{\text{Total points earned}}
{\text{Number of instances} \times 2}
\times 100.
\]

Thus, Average Score reflects occurrence-level accuracy, whereas Exact Match
requires both occurrences in an instance to be classified correctly.

\section{Hyperparameter Settings}
\label{app:hyperparams}

Table~\ref{tab:hyperparams} lists the final QLoRA fine-tuning configuration,
applied without modification across all fine-tuned models. We arrived at these
values by monitoring validation loss: an initial higher-capacity configuration
(rank 32, $\alpha=64$, learning rate $2\times10^{-4}$, dropout 0.05, weight
decay 0.01) overfit within three epochs, so we reduced adapter capacity and
learning rate and increased regularisation until the validation curve
stabilised.

\begin{table}[h]
\centering
\small
\setlength{\tabcolsep}{6pt}
\begin{tabular}{@{}ll@{}}
\toprule
\textbf{Hyperparameter} & \textbf{Value} \\
\midrule
\multicolumn{2}{@{}l}{\textit{Quantisation}} \\
Precision              & 4-bit NF4 \\
Double quantisation    & Yes \\
Compute dtype          & fp16 \\
\midrule
\multicolumn{2}{@{}l}{\textit{LoRA}} \\
Rank ($r$)             & 16 \\
$\alpha$               & 32 \\
Dropout                & 0.1 \\
Target modules         & \texttt{q,k,v,o,gate,up,down} \\
\midrule
\multicolumn{2}{@{}l}{\textit{Optimisation}} \\
Learning rate          & $1\times10^{-4}$ \\
LR schedule            & cosine \\
Warm-up ratio          & 0.1 \\
Weight decay           & 0.05 \\
Optimiser              & paged AdamW (8-bit) \\
Max gradient norm      & 0.3 \\
\midrule
\multicolumn{2}{@{}l}{\textit{Training}} \\
Effective batch size   & 16 (grad.\ accum.\ 8) \\
Max epochs             & 8 \\
Early stopping         & val.\ loss, patience 3 \\
Checkpoint             & best val.\ loss \\
Max sequence length    & 768 \\
\midrule
\multicolumn{2}{@{}l}{\textit{Hardware}} \\
GPU                    & 1$\times$ NVIDIA T4 (16\,GB) \\
Decoding (inference)   & greedy \\
\bottomrule
\end{tabular}
\caption{Final QLoRA-KD fine-tuning hyperparameters, applied identically
across Qwen2.5-1.5B, Llama-3.2-3B, and TituLLM-1B.}
\label{tab:hyperparams}
\end{table}

\section{Per-Label Analysis}
\label{sec:results-perlabel}

\begin{table*}[t]
\centering
\resizebox{\textwidth}{!}{%
\begin{tabular}{@{}l ccc c ccc c ccc@{}}
\toprule
& \multicolumn{3}{c}{\textbf{Qwen2.5-1.5B}} & 
& \multicolumn{3}{c}{\textbf{Llama-3.2-3B}} & 
& \multicolumn{3}{c}{\textbf{TituLLM-1B}} \\
\cmidrule(lr){2-4}\cmidrule(lr){6-8}\cmidrule(lr){10-12}
\textbf{Label} & P & R & F1 & & P & R & F1 & & P & R & F1 \\
\midrule
Name             & 1.00 & 1.00 & \textbf{1.00} & & 0.99 & 0.97 & \textbf{0.98} & & 1.00 & 0.97 & \textbf{0.99} \\
Concept          & 0.92 & 0.91 & 0.91 & & 0.97 & 0.85 & 0.91 & & 0.98 & 0.91 & 0.94 \\
Emotional State  & 0.97 & 0.97 & 0.97 & & 0.80 & 0.95 & 0.87 & & 0.89 & 1.00 & 0.94 \\
Collective State & 0.96 & 1.00 & 0.98 & & 0.76 & 0.96 & 0.85 & & 0.92 & 1.00 & 0.96 \\
Emotion          & 0.61 & 0.85 & 0.71 & & 0.65 & 0.55 & 0.59 & & 0.93 & 0.74 & 0.82 \\
Spiritual        & 0.89 & 0.47 & 0.62 & & 0.77 & 0.94 & 0.85 & & 0.57 & 0.89 & 0.70 \\
\bottomrule
\end{tabular}%
}
\caption{Per-label precision (P), recall (R), and F1 for the three fine-tuned
(QLoRA-KD) models. \textsc{Name} achieves near-perfect F1 across all models;
the rarest labels (\textsc{Emotion}, \textsc{Spiritual}) remain hardest.}
\label{tab:perlabel-f1}
\end{table*}
To locate where residual errors concentrate after fine-tuning, we report
per-label precision, recall, and F1 for the three QLoRA-KD models in
Table~\ref{tab:perlabel-f1}. Two patterns are consistent across all models.
First, \textsc{Name} is recovered almost perfectly (F1 $\geq 0.98$
everywhere), confirming at the label level that the dominant-meaning bias
central to our study is eliminated by distillation models no longer default
away from the name reading. Second, the errors that remain fall almost
entirely on the two rarest labels, \textsc{Emotion} and \textsc{Spiritual}
(F1 as low as 0.59 and 0.62), while the more frequent categories
(\textsc{Name}, \textsc{Concept}, \textsc{Emotional State}) are handled
reliably. This concentration of error in the long tail accounts for the gap
between Micro- and Macro-averaged F1, and indicates that the remaining
challenge is one of data scarcity in culturally rarer readings rather than a
limitation of the method itself.

\end{document}